\documentclass[11pt,letterpaper]{article}
\usepackage[hyperref]{acl2018}
\usepackage{times}
\usepackage{latexsym}
\usepackage{url}
\usepackage{hhline}
\usepackage{bm}
\usepackage{bbm} 
\usepackage{graphicx}  
\usepackage{tabularx}
\usepackage{multirow}
\usepackage{booktabs} 
\usepackage{mathrsfs}
\usepackage{tabulary}
\usepackage{makecell}
\usepackage{amsmath,amsfonts,amssymb}
\usepackage[ruled,linesnumbered]{algorithm2e}  

\usepackage{algorithmic}

\aclfinalcopy 
\def\aclpaperid{552} 

\title{Discourse Marker Augmented Network with Reinforcement Learning for Natural Language Inference}

\author{Boyuan Pan$^\dag$, Yazheng Yang$^\ddag$, Zhou Zhao$^\ddag$, Yueting Zhuang$^\ddag$, Deng Cai$^{\dag \sharp}$\thanks{corresponding author}, Xiaofei He$^{\star \dag}$\\
	$^\dag$State Key Lab of CAD$\&$CG, Zhejiang University, Hangzhou, China\\
	$^\ddag$College of Computer Science, Zhejiang University, Hangzhou, China\\
	$^\sharp$Alibaba-Zhejiang University Joint Institute of Frontier Technologies\\
	$^\star$Fabu Inc., Hangzhou, China\\
	{\tt $\{$panby, yazheng\_yang, zhaozhou, yzhuang$\}$@zju.edu.cn}\\
	{\tt $\{$dengcai, xiaofeihe$\}$@cad.zju.edu.cn} \\
}

\begin{document}
\maketitle

\begin{abstract}
Natural Language Inference (NLI), also known as Recognizing Textual Entailment (RTE), is one of the most important problems in natural language processing. It requires to infer the logical relationship between two given sentences. While current approaches mostly focus on the interaction architectures of the sentences, in this paper, we propose to transfer knowledge from some important discourse markers to augment the quality of the NLI model. We observe that people usually use some discourse markers such as ``so" or ``but" to represent the logical relationship between two sentences. These words potentially have deep connections with the meanings of the sentences, thus can be utilized to help improve the representations of them. Moreover, we use reinforcement learning to optimize a new objective function with a reward defined by the property of the NLI datasets to make full use of the labels information. Experiments show that our method achieves the state-of-the-art performance on several large-scale datasets.
\end{abstract}

\section{Introduction}
In this paper, we focus on the task of Natural Language Inference (NLI), which is known as a significant yet challenging task for natural language understanding. In this task, we are given two sentences which are respectively called \textbf{premise} and \textbf{hypothesis}. The goal is to determine whether the logical relationship between them is \emph{entailment}, \emph{neutral}, or \emph{contradiction}.

Recently, performance on NLI\cite{chen2017natural,gong2018natural,chen2017enhanced} has been significantly boosted since the release of some high quality large-scale benchmark datasets such as SNLI\cite{bowman2015large} and MultiNLI\cite{williams2017broad}. Table \ref{tab1} shows some examples in SNLI. Most state-of-the-art works focus on the interaction architectures between the premise and the hypothesis, while they rarely concerned the discourse relations of the sentences, which is a core issue in natural language understanding.

\begin{table}
	\begin{center}
		\begin{tabular}{l}
			\toprule
			\textbf{Premise}: A soccer game with multiple males\\ playing.\\
			{\textbf{Hypothesis}: Some men are playing a sport.}\\ 
			\textbf{Label}: \emph{Entailment}\\ \hline
			{\textbf{Premise}: An older and younger man smiling.}\\
			\textbf{Hypothesis}: Two men are smiling and laughing\\ at the cats playing on the floor.\\
			\textbf{Label}: \emph{Neutral}\\ \hline
            \textbf{Premise}: A black race car starts up in front of\\ a crowd of people\\
			\textbf{Hypothesis}: A man is driving down a lonely\\ road.\\
			\textbf{Label}: \emph{Contradiction}\\
			\bottomrule
		\end{tabular}
	\end{center}
	\caption{\label{tab1} Three examples in SNLI dataset.}       
\end{table}

People usually use some certain set of words to express the discourse relation between two sentences\footnote{Here sentences mean either the whole sentences or the main clauses of a compound sentence.}. These words, such as ``but" or ``and", are denoted as \emph{discourse markers}. These discourse markers have deep connections with the intrinsic relations of two sentences and intuitively correspond to the intent of NLI, such as ``but" to ``contradiction", ``so" to ``entailment", \emph{etc}. 

Very few NLI works utilize this information revealed by discourse markers. \citet{nie2017dissent} proposed to use discourse markers to help represent the meanings of the sentences. However, they represent each sentence by a single vector and directly concatenate them to predict the answer, which is too simple and not ideal for the large-scale datasets.

In this paper, we propose a \emph{Discourse Marker Augmented Network} for natural language inference, where we transfer the knowledge from the existing supervised task: Discourse Marker Prediction (DMP)\cite{nie2017dissent}, to an integrated NLI model. We first propose a sentence encoder model that learns the representations of the sentences from the DMP task and then inject the encoder to the NLI network. Moreover, because our NLI datasets are manually annotated, each example from the datasets might get several different labels from the annotators although they will finally come to a consensus and also provide a certain label. In consideration of that different confidence level of the final labels should be discriminated, we employ reinforcement learning with a reward defined by the uniformity extent of the original labels to train the model. The contributions of this paper can be summarized as follows.
\begin{itemize}
	\item Unlike previous studies, we solve the task of the natural language inference via transferring knowledge from another supervised task. We propose the Discourse Marker Augmented Network to combine the learned encoder of the sentences with the integrated NLI model.
	
	\item According to the property of the datasets, we incorporate reinforcement learning to optimize a new objective function to make full use of the labels' information.
	
	\item We conduct extensive experiments on two large-scale datasets to show that our method achieves better performance than other state-of-the-art solutions to the problem.
\end{itemize}

\section{Task Description}
\subsection{Natural Language Inference (NLI)}
In the natural language inference tasks, we are given a pair of sentences $(P,H)$, which respectively means the premise and hypothesis. Our goal is to judge whether their logical relationship between their meanings by picking a label from a small set: \emph{entailment} (The hypothesis is definitely a true description of the premise), \emph{neutral} (The hypothesis might be a true description of the premise), and \emph{contradiction} (The hypothesis is definitely a false description of the premise).

\subsection{Discourse Marker Prediction (DMP)}
For DMP, we are given a pair of sentences $(S_1,S_2)$, which is originally the first half and second half of a complete sentence. The model must predict which discourse marker was used by the author to link the two ideas from a set of candidates.

\section{Sentence Encoder Model}
\label{sec3}
Following \cite{nie2017dissent,kiros2015skip}, we use BookCorpus\cite{zhu2015aligning} as our training data for discourse marker prediction, which is a dataset of text from unpublished novels, and it is large enough to avoid bias towards any particular domain or application. After preprocessing, we obtain a dataset with the form $(S_1,S_2,m)$, which means the first half sentence, the last half sentence, and the discourse marker that connected them in the original text. Our goal is to predict the $m$ given $S_1$ and $S_2$.

We first use \emph{Glove}\cite{pennington2014glove} to transform $\{ S_{t} \}^{2}_{t=1}$ into vectors word by word and subsequently input them to a bi-directional LSTM:
\begin{equation}
\begin{aligned}
\overrightarrow{\mathbf{h}^i_t} = \overrightarrow{{\rm LSTM}} (Glove(S^i_t)), i= 1,...,|S_t| \\
\overleftarrow{\mathbf{h}^i_t} = \overleftarrow{{\rm LSTM}} (Glove(S^i_t)), i= |S_t|,...,1
\end{aligned}
\end{equation}

\begin{figure*}[t]
	\begin{center}
		\includegraphics[height=0.4 \textwidth]{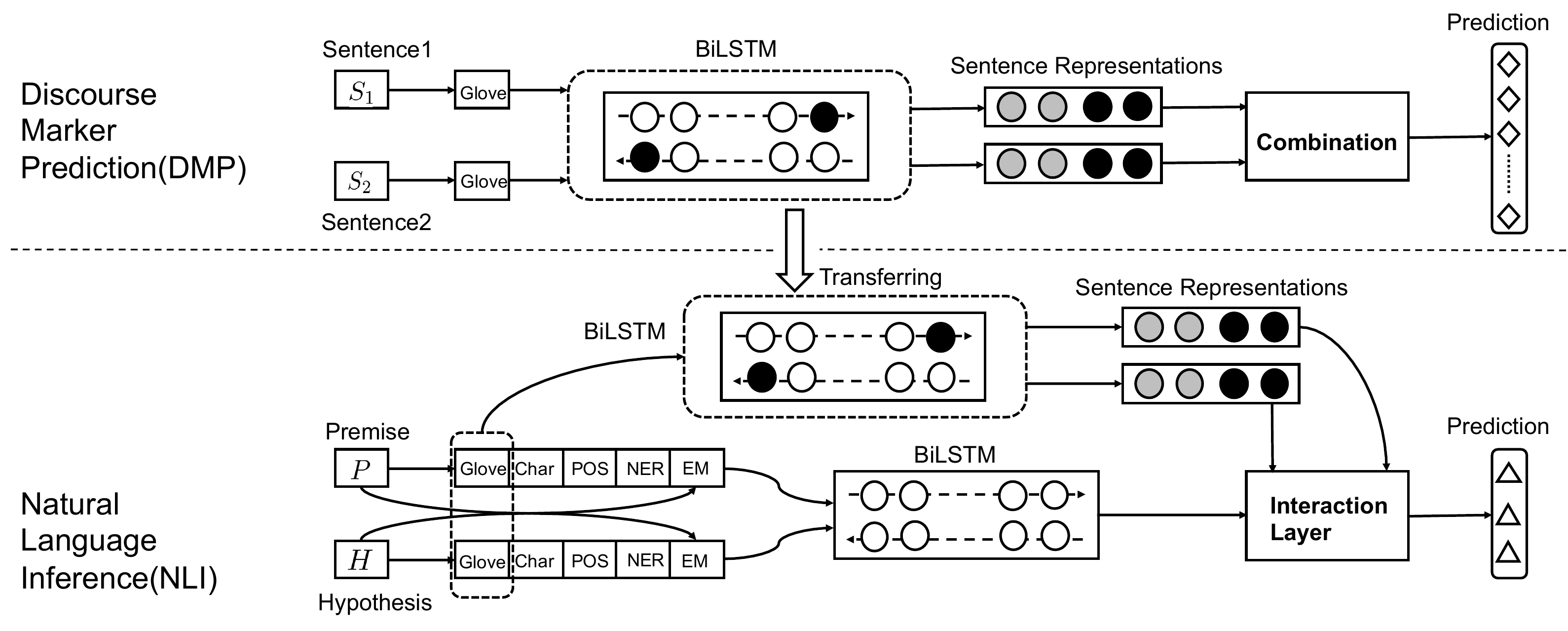}
		\caption{\label{fig1} Overview of our Discource Marker Augmented Network, comprising the part of Discourse Marker Prediction (upper) for pre-training and Natural Language Inferance (bottom) to which the learned knowledge will be transferred.}
	\end{center}
\end{figure*}

where $Glove(w)$ is the embedding vector of the word $w$ from the \emph{Glove} lookup table, $|S_t|$ is the length of the sentence $S_t$. We apply max pooling on the concatenation of the hidden states from both directions, which provides regularization and shorter back-propagation paths\cite{collobert2008unified}, to extract the features of the whole sequences of vectors:
\begin{equation}
\begin{aligned}
\overrightarrow{\mathbf{r}_t} ={\rm Max}_{dim}([\overrightarrow{\mathbf{h}^1_t};\overrightarrow{\mathbf{h}^2_t};...;\overrightarrow{\mathbf{h}^{|S_t|}_t}])\\
\overleftarrow{\mathbf{r}_t} ={\rm Max}_{dim}([\overleftarrow{\mathbf{h}^1_t};\overleftarrow{\mathbf{h}^2_t};...;\overleftarrow{\mathbf{h}^{|S_t|}_t}])\\
\end{aligned}
\end{equation}
where ${\rm Max}_{dim}$ means that the max pooling is performed across each dimension of the concatenated vectors, $[;]$ denotes concatenation. Moreover, we combine the last hidden state from both directions and the results of max pooling to represent our sentences:
\begin{equation}
\label{eq3}
\begin{aligned}
\mathbf{r}_t = [\overrightarrow{\mathbf{r}_t};\overleftarrow{\mathbf{r}_t};\overrightarrow{\mathbf{h}^{|S_t|}_t};\overleftarrow{\mathbf{h}^1_t}]
\end{aligned}
\end{equation}
where $\mathbf{r}_t$ is the representation vector of the sentence $S_t$. To predict the discource marker between $S_1$ and $S_2$, we combine the representations of them with some linear operation:
\begin{equation}
\begin{aligned}
\mathbf{r} = [\mathbf{r}_1; \mathbf{r}_2; \mathbf{r}_1 + \mathbf{r}_2 ; \mathbf{r}_1 \odot  \mathbf{r}_2]
\end{aligned}
\end{equation}
where $\odot$ is elementwise product. Finally we project $\mathbf{r}$ to a vector of label size (the total number of discourse markers in the dataset) and use softmax function to normalize the probability distribution.

\section{Discourse Marker Augmented Network}
As presented in Figure \ref{fig1}, we show how our Discourse Marker Augmented Network incorporates the learned encoder into the NLI model. 

\subsection{Encoding Layer}
We denote the premise as $P$ and the hypothesis as $H$. To encode the words, we use the concatenation of following parts:\\
\textbf{Word Embedding: }Similar to the previous section, we map each word to a vector space by using pre-trained word vectors \emph{GloVe}.\\
\textbf{Character Embedding: }We apply Convolutional Neural Networks (CNN) over the characters of each word. This approach is proved to be helpful in handling out-of-vocab (OOV) words\cite{Yang2016Words}.\\
\textbf{POS and NER tags: }We use the part-of-speech (POS) tags and named-entity recognition (NER) tags to get syntactic information and entity label of the words. Following \cite{pan2017memen}, we apply the skip-gram model\cite{mikolov2013distributed} to train two new lookup tables of POS tags and NER tags respectively. Each word can get its own POS embedding and NER embedding by these lookup tables. This approach represents much better geometrical features than common used one-hot vectors.\\
\textbf{Exact Match: }Inspired by the machine comprehension tasks\cite{Chen2017Reading}, we want to know whether every word in $P$ is in $H$ (and $H$ in $P$). We use three binary features to indicate whether the word can be exactly matched to any question word, which respectively means original form, lowercase and lemma form.

For encoding, we pass all sequences of vectors into a bi-directional LSTM and obtain:
\begin{equation}
\begin{aligned}
&\mathbf{p}_i = {\rm BiLSTM} (f_{rep}(P_i),\mathbf{p}_{i-1}), i= 1,...,n\\
&\mathbf{u}_j = {\rm BiLSTM} (f_{rep}(H_j),\mathbf{u}_{j-1}), j= 1,...,m
\end{aligned}
\end{equation}
where $f_{rep}(\mathbf{x}) = [{\rm Glove}(\mathbf{x}); {\rm Char}(\mathbf{x}); {\rm POS}(\mathbf{x});$ $ {\rm NER}(\mathbf{x}); {\rm EM}(\mathbf{x})]$ is the concatenation of the embedding vectors and the feature vectors of the word $\mathbf{x}$, $n = |P|$, $m = |H|$.

\subsection{Interaction Layer}
In this section, we feed the results of the encoding layer and the learned sentence encoder into the attention mechanism, which is responsible for linking and fusing information from the premise and the hypothesis words. 

We first obtain a similarity matrix $\mathbf{A} \in R^{n \times m}$ between the premise and hypothesis by 
\begin{equation}
\label{eq6}
\begin{aligned}
\mathbf{A}_{ij} = \mathbf{v}_{1}^{\top}[\mathbf{p}_i;\mathbf{u}_j;\mathbf{p}_i \circ \mathbf{u}_j;\mathbf{r}_p; \mathbf{r}_h]
\end{aligned}
\end{equation}
where $\mathbf{v}_{1}$ is the trainable parameter, $\mathbf{r}_p$ and $\mathbf{r}_h$ are sentences representations from the equation (\ref{eq3}) learned in the Section \ref{sec3}, which denote the premise and hypothesis respectively. In addition to previous popular similarity matrix, we incorporate the relevance of each word of $P(H)$ to the whole sentence of $H(P)$. Now we use $\mathbf{A}$ to obtain the attentions and the attended vectors in both directions.

To signify the attention of the $i$-th word of $P$ to every word of $H$, we use the weighted sum of $\mathbf{u}_j$ by $\mathbf{A}_{i:}$:
\begin{equation}
\label{eq7}
\begin{aligned}
\tilde{\mathbf{u}}_i = \sum_j \mathbf{A}_{ij} \cdot \mathbf{u}_j
\end{aligned}
\end{equation}
where $\tilde{\mathbf{u}}_i$ is the attention vector of the $i$-th word of $P$ for the entire $H$. In the same way, the $\tilde{\mathbf{p}}_j$ is obtained via:
\begin{equation}
\begin{aligned}
\tilde{\mathbf{p}}_j = \sum_i \mathbf{A}_{ij} \cdot \mathbf{p}_i
\end{aligned}
\end{equation}

To model the local inference between aligned word pairs, we integrate the attention vectors with the representation vectors via:
\begin{equation}
\begin{aligned}
\hat{\mathbf{p}}_i &= f([\mathbf{p}_i;\tilde{\mathbf{u}}_i; \mathbf{p}_i - \tilde{\mathbf{u}}_i; \mathbf{p}_i \odot \tilde{\mathbf{u}}_i])\\
\hat{\mathbf{u}}_j &= f([\mathbf{u}_j;\tilde{\mathbf{p}}_j; \mathbf{u}_j - \tilde{\mathbf{p}}_j; \mathbf{u}_j \odot \tilde{\mathbf{p}}_j])
\end{aligned}
\end{equation}
where $f$ is a 1-layer feed-forward neural network with the ReLU activation function, $\hat{\mathbf{p}}_i$ and $\hat{\mathbf{u}}_j$ are local inference vectors. Inspired by \cite{seo2016bidirectional} and \cite{chen2017natural}, we use a modeling layer to capture the interaction between the premise and the hypothesis. Specifically, we use bi-directional LSTMs as building blocks:
\begin{equation}
\begin{aligned}
&\mathbf{p}^{M}_i = {\rm BiLSTM} (\hat{\mathbf{p}}_i,\mathbf{p}^{M}_{i-1})\\
&\mathbf{u}^{M}_j = {\rm BiLSTM} (\hat{\mathbf{u}}_j,\mathbf{u}^{M}_{j-1})
\end{aligned}
\end{equation}
Here, $\mathbf{p}^{M}_i$ and $\mathbf{u}^{M}_j$ are the modeling vectors which contain the crucial information and relationship among the sentences.

We compute the representation of the whole sentence by the weighted average of each word:
\begin{equation}
\begin{aligned}
\mathbf{p}^M = \sum_i \frac{{\rm exp}(\mathbf{v}_2^{\top} \mathbf{p}_i^{M})}{\sum_{i'}{\rm exp}(\mathbf{v}_2^{\top} \mathbf{p}_{i'}^{M})}\mathbf{p}_i^{M}\\
\mathbf{u}^M = \sum_j \frac{{\rm exp}(\mathbf{v}_3^{\top} \mathbf{u}^M_j)}{\sum_{j'}{\rm exp}(\mathbf{v}_3^{\top} \mathbf{u}^M_{j'})}\mathbf{u}^M_j
\end{aligned}
\end{equation}
where $\mathbf{v}_2,\mathbf{v}_3$ are trainable vectors. We don't share these parameter vectors in this seemingly parallel strucuture because there is some subtle difference between the premise and hypothesis, which will be discussed later in Section \ref{sec5}.

\subsection{Output Layer}
The NLI task requires the model to predict the logical relation from the given set: \emph{entailment}, \emph{neutral} or \emph{contradiction}. We obtain the probability distribution by a linear function with softmax function:
\begin{equation}
\label{eq12}
\begin{aligned}
\mathbf{d} = {\rm softmax}(\mathbf{W}[\mathbf{p}^M; \mathbf{u}^M ; \mathbf{p}^M \odot \mathbf{u}^M ; \mathbf{r}_p \odot \mathbf{r}_h])
\end{aligned}
\end{equation}
where $\mathbf{W}$ is a trainable parameter. We combine the representations of the sentences computed above with the representations learned from DMP to obtain the final prediction.

 \begin{table}[t]
	\centering
	\begin{tabular}{c|lp{1cm}lp{1cm}lp{1cm}lp{1cm}lp{1cm}}
		\hline
		\multicolumn{1}{c|}{\textbf{Label}} & \multicolumn{2}{c}{\textbf{SNLI}} &  \multicolumn{2}{c}{\textbf{MultiNLI}} \\ \cline{2-5} 
		\multicolumn{1}{c|}{\textbf{Number}}                        & \textbf{Correct}          & \textbf{Total}    & \textbf{Correct}     & \textbf{Total}      \\ \hline
		1    &  510711 & 510711 & 392702  &  392702     \\
		2   & 0 & 0    & 0   &  0 \\
		3 & 8748 & 0     & 3045  &  0  \\
		4 &  16395 & 2199  & 4859  & 0 \\ 
		5    &  33179 & 56123 & 11743 &  19647     \\
		\hline
	\end{tabular}
	\vspace{2mm}
	\caption{Statistics of the labels of SNLI and MuliNLI. \textbf{Total} means the number of examples whose number of annotators is in the left column. \textbf{Correct} means the number of examples whose number of correct labels from the annotators is in the left column. }     
	\label{tab2}  
\end{table}

\subsection{Training}
As shown in Table \ref{tab2}, many examples from our datasets are labeled by several people, and the choices of the annotators are not always consistent. For instance, when the label number is 3 in SNLI, ``total=0" means that no examples have 3 annotators (maybe more or less); ``correct=8748" means that there are 8748 examples whose number of correct labels is 3 (the number of annotators maybe 4 or 5, but some provided wrong labels). Although all the labels for each example will be unified to a final (correct) label, diversity of the labels for a single example indicates the low confidence of the result, which is not ideal to only use the final label to optimize the model. 

We propose a new objective function that combines both the log probabilities of the ground-truth label and a reward defined by the property of the datasets for the reinforcement learning. The most widely used objective function for the natural language inference is to minimize the negative log cross-entropy loss:
\begin{equation}
\begin{aligned}
J_{CE}(\Theta) = - \frac{1}{N} \sum_k^N log(\mathbf{d}_l^k) 
\end{aligned}
\end{equation}
where $\Theta$ are all the parameters to optimize, $N$ is the number of examples in the dataset, $\mathbf{d}_l$ is the probability of the ground-truth label $l$.

However, directly using the final label to train the model might be difficult in some situations, where the example is confusing and the labels from the annotators are different. For instance, consider an example from the SNLI dataset:
\begin{itemize}
\item $P$: ``A smiling costumed woman is holding an umbrella." 
\item  $H$: ``A happy woman in a fairy costume holds an umbrella." 
\end{itemize}
 The final label is \emph{neutral}, but the original labels from the five annotators are \emph{neural}, \emph{neural}, \emph{entailment}, \emph{contradiction}, \emph{neural}, in which case the relation between ``smiling" and ``happy" might be under different comprehension. The final label's confidence of this example is obviously lower than an example that all of its labels are the same. To simulate the thought of human being more closely, in this paper, we tackle this problem by using the REINFORCE algorithm\cite{williams1992simple} to minimize the negative expected reward, which is defined as:
\begin{equation}
\begin{aligned}
J_{RL}(\Theta) = - \mathbb{E}_{l \sim \pi (l | P, H)}[R(l, \{l^*\})]
\end{aligned}
\end{equation}
where $\pi (l | P, H)$ is the previous action policy that predicts the label given $P$ and $H$, $\{l^*\}$ is the set of annotated labels, and
\begin{equation}
\begin{aligned}
R(l, \{l^*\}) = \frac{{\rm number~of}~l~{\rm in}~ \{l^*\}}{ |\{l^*\}|}
\end{aligned}
\end{equation}
is the reward function defined to measure the distance to all the ideas of the annotators.

To avoid of overwriting its earlier results and further stabilize training, we use a linear function to integrate the above two objective functions:
\begin{equation}
\begin{aligned}
J(\Theta) = \lambda J_{CE}(\Theta) + (1- \lambda)J_{RL}(\Theta)
\end{aligned}
\end{equation}
where $\lambda$ is a tunable hyperparameter.

\section{Experiments}
\label{sec5}
\subsection{Datasets}
\textbf{BookCorpus: }We use the dataset from BookCorpus\cite{zhu2015aligning} to pre-train our sentence encoder model. We preprocessed and collected discourse markers from BookCorpus as \cite{nie2017dissent}. We finally curated a dataset of 6527128 pairs of sentences for 8 discourse markers, whose statistics are shown in Table \ref{tab3}.\\
\textbf{SNLI: }Stanford Natural Language Inference\cite{bowman2015large} is a collection of more than 570k  human annotated sentence pairs labeled for entailment, contradiction, and semantic independence. SNLI is two orders of magnitude larger than all other resources of its type. The premise data is extracted from the captions of the Flickr30k corpus\cite{young2014image}, the hypothesis data and the labels are manually annotated. The original SNLI corpus contains also “the other” category, which includes the sentence pairs lacking consensus among multiple human annotators. We remove this category and use the same split as in \cite{bowman2015large} and other previous work.\\
\textbf{MultiNLI: } Multi-Genre Natural Language Inference\cite{williams2017broad} is another large-scale corpus for the task of NLI. MultiNLI has 433k sentences pairs and is in the same format as SNLI, but it includes a more diverse range of text, as well as an auxiliary test set for cross-genre transfer evaluation. Half of these selected genres appear in training set while the rest are not, creating in-domain (matched) and cross-domain (mismatched) development/test sets. 

\begin{table}[t]
	\begin{center}
		\begin{tabular}{c|c}
			\toprule
			\textbf{Discourse Marker} &  \textbf{Percentage(\%)} \\
			\midrule
			\emph{but}  & 57.12 \\ 
			\emph{because}  & 9.41 \\ 
			\emph{if}  & 29.78 \\ 
			\emph{when}  & 25.32 \\ 
			\emph{so}  & 31.01 \\ 
			\emph{although}  & 1.76 \\ 
			\emph{before}  & 15.52 \\ 
			\emph{still}  & 11.29 \\ 
			\bottomrule
		\end{tabular}
		\vspace{2mm}
	\end{center}
	\caption{\label{tab3} Statistics of discouse markers in our dataset from BookCorpus.}       
\end{table}

\begin{table*}[t]
	\begin{center}
		\begin{tabular}{lp{1cm}lp{2cm}cp{2cm}c}
			\toprule
			\multicolumn{1}{c}{\multirow{2}{*}{\textbf{Method}}} & \multicolumn{1}{c}{\multirow{2}{*}{\textbf{SNLI}}} &  \multicolumn{2}{c}{\textbf{MultiNLI}} \\ \cline{3-4} 
			\multicolumn{1}{c}{}  & \multicolumn{1}{c}{}  &		
			\textbf{Matched} &  \textbf{Mismatched} \\
			\midrule
			300D LSTM encoders\cite{bowman2016fast}  & 80.6 & -- & --\\ 
			300D Tree-based CNN encoders\cite{mou2016natural} & 82.1 & -- & --\\
			4096D BiLSTM with max-pooling\cite{conneau2017supervised} & 84.5 & -- & --\\
			600D Gumbel TreeLSTM encoders\cite{choi2017learning}  & 86.0  & -- & --\\ 
			600D Residual stacked encoders\cite{nie2017shortcut}& 86.0  & 74.6 & 73.6 \\ \hline
			Gated-Att BiLSTM\cite{chen2017recurrent} & -- & 73.2 & 73.6\\
			100D LSTMs with attention\cite{rocktaschel2015reasoning} & 83.5 & -- & --\\
			300D re-read LSTM\cite{sha2016reading} & 87.5& -- & --\\
			DIIN\cite{gong2018natural}  & 88.0 & 78.8 & 77.8\\
			Biattentive Classification Network\cite{mccann2017learned} &88.1 & -- & --\\
			300D CAFE\cite{tay2017compare} & 88.5 & 78.7 & 77.9\\
			KIM\cite{chen2017natural} & 88.6& -- & -- \\
			600D ESIM + 300D Syntactic TreeLSTM\cite{chen2017enhanced}  & 88.6 & -- & --\\ 
			\textbf{DMAN} & \textbf{88.8}  & \textbf{78.9} & \textbf{78.2}\\ \hline
			BiMPM(Ensemble)\cite{wang2017bilateral}  & 88.8 & -- & --\\ 
			DIIN(Ensemble)\cite{gong2018natural}  & 88.9 & 80.0 & 78.7 \\ 
			KIM(Ensemble)\cite{chen2017natural}  & 89.1 & -- & --\\ 
			300D CAFE(Ensemble)\cite{tay2017compare}  & 89.3 & 80.2 & 79.0 \\ 
			\textbf{DMAN}(Ensemble) & \textbf{89.6} & \textbf{80.3} & \textbf{79.4} \\
			\bottomrule
		\end{tabular}
		\vspace{2mm}
	\end{center}
	\caption{\label{tab4} Performance on the SNLI dataset and the MultiNLI dataset. In the top part, we show sentence encoding-based models; In the medium part, we present the performance of integrated neural network models; In the bottom part, we show the results of ensemble models.}       
\end{table*}

\subsection{Implementation Details}
We use the Stanford CoreNLP toolkit\cite{manning2014stanford} to tokenize the words and generate POS and NER tags. The word embeddings are initialized by 300d \emph{Glove}\citep{pennington2014glove}, the dimensions of POS and NER embeddings are 30 and 10. The dataset we use to train the embeddings of POS tags and NER tags are the training set given by SNLI. We apply Tensorflow r1.3 as our neural network framework. We set the hidden size as 300 for all the LSTM layers and apply dropout\cite{srivastava2014dropout} between layers with an initial ratio of 0.9, the decay rate as 0.97 for every 5000 step. We use the AdaDelta for optimization as described in \cite{zeiler2012adadelta} with $\rho$ as 0.95 and $\epsilon$ as 1e-8. We set our batch size as 36 and the initial learning rate as 0.6. The parameter $\lambda$ in the objective function is set to be 0.2. For DMP task, we use stochastic gradient descent with initial learning rate as 0.1, and we anneal by half each time the validation accuracy is lower than the previous epoch. The number of epochs is set to be 10, and the feedforward dropout rate is 0.2. The learned encoder in subsequent NLI task is trainable. Code is available at \url{https://github.com/ZJULearning/DMP}.

\subsection{Results}
In table \ref{tab4}, we compare our model to other competitive published models on SNLI and MultiNLI. As we can see, our method Discourse Marker Augmented Network (DMAN) clearly outperforms all the baselines and achieves the state-of-the-art results on both datasets.

The methods in the top part of the table are sentence encoding based models. \citet{bowman2016fast} proposed a simple baseline that uses LSTM to encode the whole sentences and feed them into a MLP classifier to predict the final inference relationship, they achieve an accuracy of 80.6\% on SNLI. \citet{nie2017shortcut} test their model on both SNLI and MiltiNLI, and achieves competitive results. 

In the medium part, we show the results of other neural network models. Obviously, the performance of most of the integrated methods are better than the sentence encoding based models above. Both DIIN\cite{gong2018natural} and CAFE\cite{tay2017compare} exceed other methods by more than 4\% on MultiNLI dataset. However, our DMAN achieves 88.8\% on SNLI, 78.9\% on matched MultiNLI and 78.2\% on mismatched MultiNLI, which are all best results among the baselines.

We present the ensemble results on both datasets in the bottom part of the table \ref{tab4}. We build an ensemble model which consists of 10 single models with the same architecture but initialized with different parameters. The performance of our model achieves 89.6\% on SNLI, 80.3\% on matched MultiNLI and 79.4\% on mismatched MultiNLI, which are all state-of-the-art results.

\begin{table}
	\begin{tabular}{p{5.3cm}c}
		\toprule
		\textbf{Ablation Model}	&  \textbf{Accuracy} \\
		\midrule
		Only Sentence Encoder Model & 83.37 \\
		No Sentence Encoder Model & 87.24 \\ 
		No Char Embedding & 87.95 \\ 
		No POS Embedding & 88.76 \\ 
		No NER Embedding & 88.71 \\ 
		No Exact Match & 88.26 \\
		No REINFORCE & 88.41 \\ \hline
		DMAN     & 88.83     \\
		\bottomrule
	\end{tabular}
	\vspace{2mm}
	\caption{Ablations on the SNLI development dataset.}       
	\label{tab5}
\end{table}

\subsection{Ablation Analysis}
As shown in Table \ref{tab5}, we conduct an ablation experiment on SNLI development dataset to evaluate the individual contribution of each component of our model. Firstly we only use the results of the sentence encoder model to predict the answer, in other words, we represent each sentence by a single vector and use dot product with a linear function to do the classification. The result is obviously not satisfactory, which indicates that only using sentence embedding from discourse markers to predict the answer is not ideal in large-scale datasets. We then remove the sentence encoder model, which means we don't use the knowledge transferred from the DMP task and thus the representations $\mathbf{r}_p$ and $\mathbf{r}_h$ are set to be zero vectors in the equation (\ref{eq6}) and the equation (\ref{eq12}). We observe that the performance drops significantly to 87.24\%, which is nearly 1.5\% to our DMAN model, which indicates that the discourse markers have deep connections with the logical relations between two sentences they links. When we remove the character-level embedding and the POS and NER features, the performance drops a lot. We conjecture that those feature tags help the model represent the words as a whole while the char-level embedding can better handle the out-of-vocab (OOV) or rare words. The exact match feature also demonstrates its effectiveness in the ablation result. Finally, we ablate the reinforcement learning part, in other words, we only use the original loss function to optimize the model (set $\lambda = 1$). The result drops about 0.5\%, which proves that it is helpful to utilize all the information from the annotators.

\begin{figure}[t]
	\includegraphics[width=0.52 \textwidth]{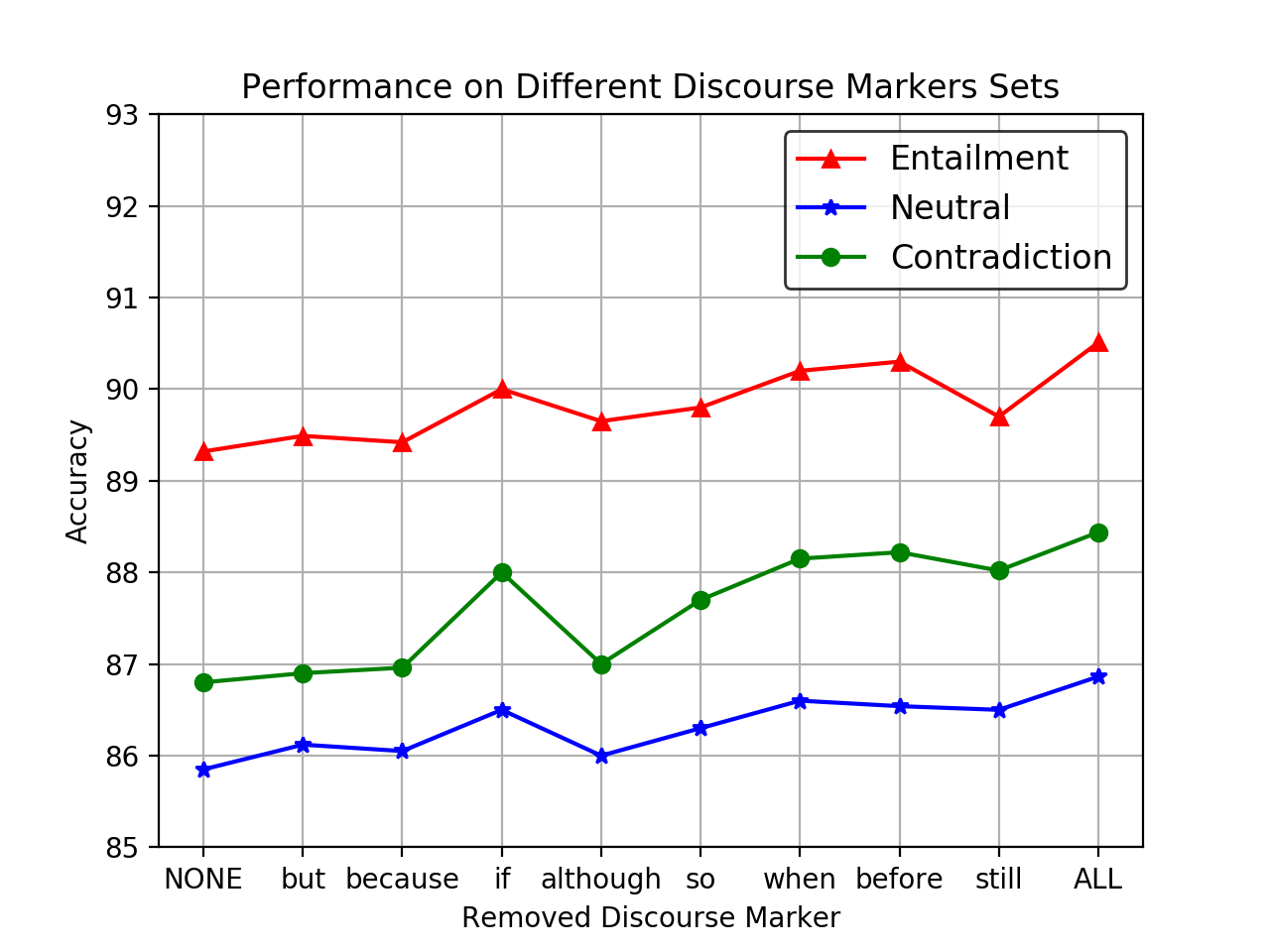}
	\caption{\label{fig2} Performance when the sentence encoder is pretrained on different discourse markers sets. ``NONE" means the model doesn't use any discourse markers; ``ALL" means the model use all the discourse markers.}
\end{figure}

\begin{figure}[t]
	\includegraphics[width=0.5 \textwidth]{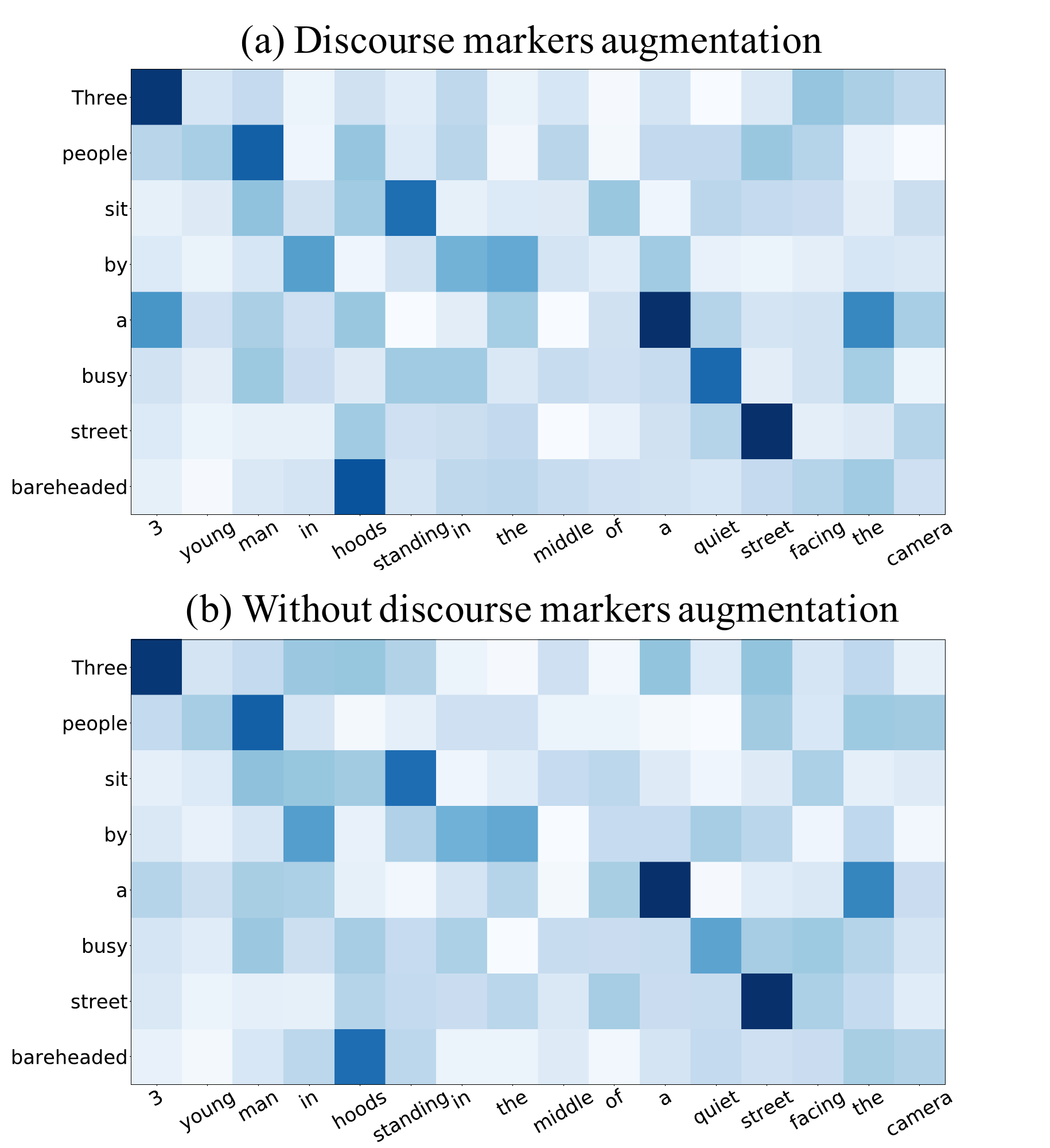}
	\caption{\label{fig3} Comparison of the visualized similarity relations.}
\end{figure}

\subsection{Semantic Analysis}
\label{sec5.5}
In Figure \ref{fig2}, we show the performance on the three relation labels when the model is pre-trained on different discourse markers sets. In other words, we removed discourse marker from the original set each time and use the rest 7 discourse markers to pre-train the sentence encoder in the DMP task and then train the DMAN. As we can see, there is a sharp decline of accuracy when removing ``but", ``because" and ``although". We can intuitively speculate that ``but" and ``although" have direct connections with the contradiction label (which drops most significantly) while ``because" has some links with the entailment label. We observe that some discourse markers such as ``if" or ``before" contribute much less than other words which have strong logical hints, although they actually improve the performance of the model. Compared to the other two categories, the ``contradiction" label examples seem to benefit the most from the pre-trained sentence encoder.

\subsection{Visualization}
 In Figure \ref{fig3}, we also provide a visualized analysis of the hidden representation from similarity matrix $\mathbf{A}$ (computed in the equation (\ref{eq6})) in the situations that whether we use the discourse markers or not. We pick a sentence pair whose premise is ``\emph{3 young man in hoods standing in the middle of a quiet street facing the camera.}" and hypothesis is ``\emph{Three people sit by a busy street bareheaded.}" We observe that the values are highly correlated among the synonyms like ``people" with ``man", ``three" with ``3" in both situations. However, words that might have contradictory meanings like ``hoods" with ``bareheaded", ``quiet" with ``busy" perform worse without the discourse markers augmentation, which conforms to the conclusion that the ``contradiction" label examples benefit a lot which is observed in the Section \ref{sec5.5}.

\section{Related Work}
\subsection{Discourse Marker Applications}
This work is inspired most directly by the DisSent model and Discourse Prediction Task of \citet{nie2017dissent}, which introduce the use of the discourse markers information for the pretraining of sentence encoders. They follow \cite{kiros2015skip} to collect a large sentence pairs corpus from BookCorpus\cite{zhu2015aligning} and propose a sentence representation based on that. They also apply their pre-trained sentence encoder to a series of natural language understanding tasks such as sentiment analysis, question-type, entailment, and relatedness. However, all those datasets are provided by \citet{conneau2017supervised} for evaluating sentence embeddings and are almost all small-scale and are not able to support more complex neural network. Moreover, they represent each sentence by a single vector and directly combine them to predict the answer, which is not able to interact among the words level.

In closely related work, \citet{jernite2017discourse} propose a model that also leverage discourse relations. However, they manually group the discourse markers into several categories based on human knowledge and predict the category instead of the explicit discourse marker phrase. However, the size of their dataset is much smaller than that in 
\cite{nie2017dissent}, and sometimes there has been disagreement among annotators about what exactly is the correct categorization of discourse relations\cite{hobbs1990literature}.

Unlike previous works, we insert the sentence encoder into an integrated network to augment the semantic representation for NLI tasks rather than directly combining the sentence embeddings to predict the relations.

\subsection{Natural Language Inference}
Earlier research on the natural language inference was based on small-scale datasets\cite{marelli2014sick}, which relied on traditional methods such as shallow methods\cite{glickman2005web}, natural logic methods\cite{maccartney2007natural}, \emph{etc}. These datasets are either not large enough to support complex deep neural network models or too easy to challenge natural language. 

Large and complicated networks have been successful in many natural language processing tasks\cite{zhu2017next,Chen2017User,pan2017keyword}. Recently, \citet{bowman2015large} released Stanford Natural language Inference (SNLI) dataset, which is a high-quality and large-scale benchmark, thus inspired many significant works\cite{bowman2016fast,mou2016natural,vendrov2016order,conneau2017supervised,wang2017bilateral,gong2018natural,mccann2017learned,chen2017natural,choi2017learning,tay2017compare}. Most of them focus on the improvement of the interaction architectures and obtain competitive results, while transfer learning from external knowledge is popular as well. \citet{vendrov2016order} incorpated Skipthought\cite{kiros2015skip}, which is an unsupervised sequence model that has been proven to generate useful sentence embedding. \citet{mccann2017learned} proposed to transfer the pre-trained encoder from the neural machine translation (NMT) to the NLI tasks.

Our method combines a pre-trained sentence encoder from the DMP task with an integrated NLI model to compose a novel framework. Furthermore, unlike previous studies, we make full use of the labels provided by the annotators and employ policy gradient to optimize a new objective function in order to simulate the thought of human being.

\section{Conclusion}
In this paper, we propose \emph{Discourse Marker Augmented Network} for the task of the natural language inference. We transfer the knowledge learned from the discourse marker prediction task to the NLI task to augment the semantic representation of the model. Moreover, we take the various views of the annotators into consideration and employ reinforcement learning to help optimize the model. The experimental evaluation shows that our model achieves the state-of-the-art results on SNLI and MultiNLI datasets. Future works involve the choice of discourse markers and some other transfer learning sources.

\section{Acknowledgements}
This work was supported in part by the National Nature Science Foundation of China (Grant Nos: 61751307), in part by the grant ZJU Research 083650 of the ZJUI Research Program from Zhejiang University and in part by the National Youth Top-notch Talent Support Program. The experiments are supported by Chengwei Yao in the Experiment Center of the College of Computer Science and Technology, Zhejiang university.

\bibliography{NLI}
\bibliographystyle{acl_natbib}

\end{document}